\documentclass[letterpaper, 10 pt, conference]{ieeeconf}  

\IEEEoverridecommandlockouts                              

\usepackage{amssymb}
\usepackage{bbm}
\usepackage{graphicx}
\usepackage{amsmath}
\usepackage{amssymb}
\usepackage{subfig}
\usepackage[table,xcdraw]{xcolor}

\usepackage{amsmath, amsfonts}
\usepackage{dsfont}
\title{\LARGE \bf
Single Object Tracking through a Fast and Effective Single-Multiple Model Convolutional Neural Network   
}

\author{Faraz Lotfi$^{1}$ and Hamid D. Taghirad$^{1}$
\thanks{*This work was not supported by any organization}
\thanks{$^{1}$Authors are with the Advanced Robotics and Automated Systems (ARAS), Industrial Control Center of
	Excellence, Faculty of Electrical Engineering, K. N. Toosi University of Technology,
	Tehran, Iran
        {\tt\small F.Lotfi@email.kntu.ac.ir, Taghirad@kntu.ac.ir}}%
}

\begin{document}

\maketitle
\thispagestyle{empty}
\pagestyle{empty}

\begin{abstract}
Object tracking becomes critical especially when similar objects are present in the same area. Recent state-of-the-art (SOTA) approaches are proposed based on taking a matching network with a heavy structure to distinguish the target from other objects in the area which indeed drastically downgrades the performance of the tracker in terms of speed. Besides, several candidates are considered and processed to localize the intended object in a region of interest for each frame which is time-consuming. In this article, a special architecture is proposed based on which in contrast to the previous approaches, it is possible to identify the object location in a single shot while taking its template into account to distinguish it from the similar objects in the same area. In brief, first of all, a window containing the object with twice the target size is considered. This window is then fed into a fully convolutional neural network (CNN) to extract a region of interest (RoI) in a form of a matrix for each of the frames. In the beginning, a template of the target is also taken as the input to the CNN. Considering this RoI matrix, the next movement of the tracker is determined based on a simple and fast method. Moreover, this matrix helps to estimate the object size which is crucial when it changes over time. Despite the absence of a matching network, the presented tracker performs comparatively with the SOTA in challenging situations while having a super speed compared to them (up to $120 FPS$ on 1080ti). To investigate this claim, a comparison study is carried out on the GOT-10k dataset. Results reveal the outstanding performance of the proposed method in fulfilling the task.
\end{abstract}

\section{Introduction}

Object tracking is a major task to be accomplished in computer vision applications. Mostly, it is utilized to take into account the information between several frames. Regarding the importance of this subject, numerous effort has been made to investigate an appropriate approach. 
Some of these methods are based on detecting objects and associating unique ids to the same objects in a sequence of frames \cite{mot_lit} while the others focus on the tracking task exclusively \cite{object_tracking}. 
Investigating the detail of the tracking-by-detection approaches, researchers commonly intend to solve the multiple object tracking problem. In this regard, \cite{tracklets} proposes to use tracklets based on a deep associate method while utilizing a tracking-by-detection paradigm. The other approach in \cite{deep_attentive} suggests a reciprocative learning algorithm while utilizing visual attention. In \cite{adversarial_}, adversarial learning is employed to find the most robust features of the targets in a form of a mask. Single object tracking, on the other hand, is of high importance for which plenty of researches are conducted in this area \cite{Lasot_}. In \cite{sot1}, a single object tracker is presented based on fuzzy least squares support vector machine. Another method is suggested based on an online fusion of trackers \cite{sot2}.  Some of the approaches are developed to deal with special problems in a tracking scenario such as occlusions \cite{robust_tracking,robust_tracking2} while others extend the tracking task to a specific application \cite{drone,aircraft,surgical_lotfi}. 

A wide research has been conducted to perform the tracking without detection. To start with, a meta-updater is proposed in \cite{LTMU} by which an online-update-based tracker is designed to accomplish the local tracking. This is done by the use of a cascaded LSTM module. This idea helps the long-term tracking to benefit from the perfect short-term trackers. Another method presented in \cite{Global} suggests a baseline for performing a global search (over a very large area). By this means, there is no need for an online learning or any changes in each of the scale and the trajectory. Besides, this tracker is mainly developed based on the two-stage detectors. In \cite{SPLT}, on the other hand, a tracking framework is presented based on the proposed skimming and perusal modules. The latter one stands for a module consisting of two main parts. Firstly, a bounding box regressor generates several candidates. Then, a verifier is applied to identify the optimal candidate while taking its confidence score into account to infer about the object presence. Note, all the mentioned trackers employ the similar approach of using a global search along with a verifier module responsible for recognizing the intended object. The verifier module is often based on ResNet-50 architecture and acts as a matching network. They also utilize the SiamRPN-based network \cite{siamrpn} to produce some region candidates for the subsequent tasks. 
On the other hand, despite the other problems related to computer vision, the object tracking task requires an online prediction of the object appearance model which should be embedded into the tracker's architecture to realize an End-to-End training. In this regard a popular approach is the Siamese paradigm \cite{Siamese1,Siamese2} where a feature template of the target is predicted regardless of the background information. In \cite{DiMP}, however, the authors suggest to consider both the target and the background appearance information to predict the required model based on defining a discriminative learning loss. Previous to this method the same backbone had been employed by \cite{atom} which is basically proposed using a ResNet-18 architecture along with an online discriminative learning.             

\begin{figure}[]
	\begin{center}
		\includegraphics[width=1\linewidth]{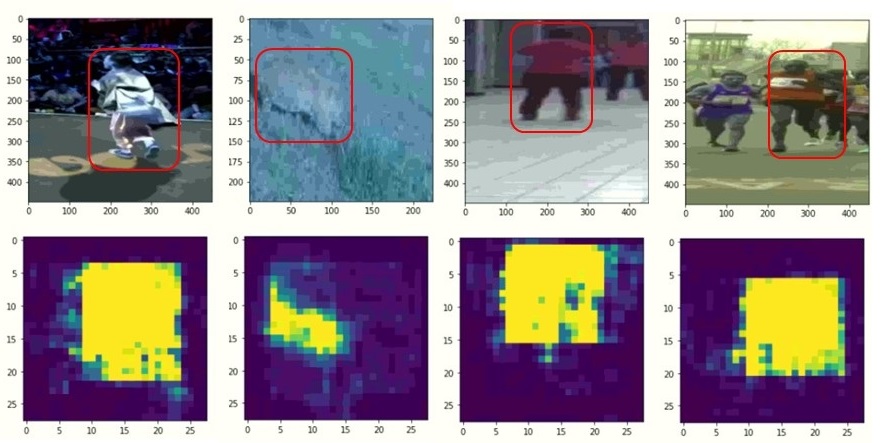}
	\end{center}
	\caption{This figure illustrates the output of the RoI extractor module when trained on GOT-10k. These four images demonstrate the desired performance of the proposed architecture in distinguishing the intended object under various situations such as cluttered background, similar objects in the RoI, etc. The effective part that realizes such a capability is the distinct branch considered with the intended object template as its input (See Fig. \ref{fig:fig1}). Note, the template is not updated through frames, and is given to the tracker just once in the first frame.}
	\label{fig:fig0}
\end{figure}

As it is seen a vital issue in object tracking is to elicit exclusive features of the target to distinguish it from both the background and the other similar objects. In this work, our focus is on presenting an appropriate architecture based on a novel approach while addressing the challenging issues mainly the presence of similar objects. 
In this article, the proposed object tracker takes the advantage of using a light CNN model along with four input branches to track an object while taking both its template and its size into account. Besides, it is suggested to take only a small part of the image containing the specified object. This part is then updated and displaced by the proposed method. Based on this approach, the mentioned region of interest is enough to track the object even in a long-term tracking scenario. The challenge, then, is to determine an appropriate new RoI for each of the incoming frames. Assuming an initial RoI with a size twice the object one, we present a new object tracker to accomplish the object tracking task. In brief, an object localizer utilizes the RoI as its input to elicit a matrix representing the object location. This module is a fully CNN with several input branches to process each object regarding its specific size while taking the intended object template from the first frame into account, as well. To clarify, in each processing, two of four branches are activated one of which is the template related branch, and the other is one of the three branches corresponding to the object size. Taking the object template branch helps to integrate the exclusive features of the target to distinguish it from similar objects. Fig. \ref{fig:fig0} presents some sample outputs of the proposed model trained on GOT-10k dataset. The methodology is presented in detail in the next section. Then experimental results are reported while comparing the performance of the proposed tracker with state-of-the-art approaches. Finally, the last section concludes the paper.                  

\begin{figure*}[]
	\begin{center}
		\includegraphics[width=0.9\linewidth]{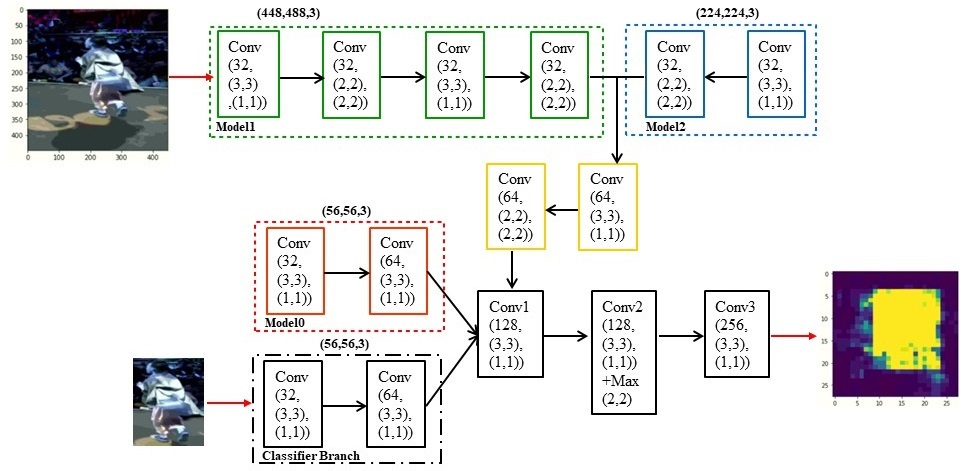}
	\end{center}
	\caption{Schematic of the architecture presented as the RoI extractor. This is partly inspired by the concept presented in \cite{us}. Compared to this reference, instead of the two output matrices, we employ the RoI matrix while considering a higher resolution of $(28,28)$. Besides, a new branch is added to integrate the target template in the object tracking.}.  
	\label{fig:fig1}
\end{figure*}

\section{Methodology}
This section focuses on the methodology of the presented approach. There are several points to present regarding the proposed method. In what follows, first, the main model used to extract the RoI matrix from an image is investigated. This model is inspired by \cite{us}, and it is shown to be applicable in obtaining an accurate object location from an ROI in a single image. Then, the other parts are presented and explained, precisely.

\begin{figure}[]
	\begin{center}
		\includegraphics[width=0.6\linewidth]{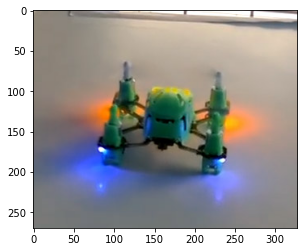}\\
		\includegraphics[width=1\linewidth]{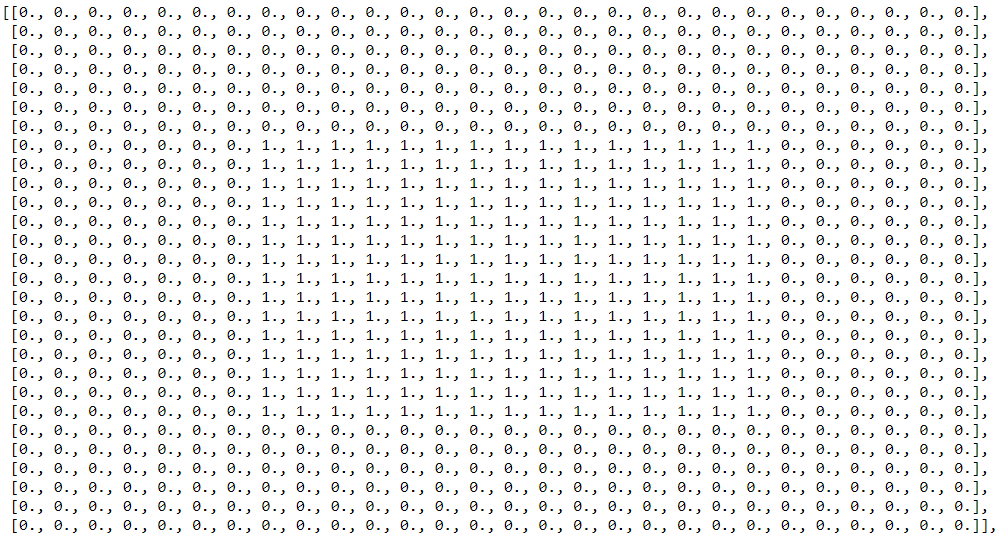}
	\end{center}
	\caption{The concept of using the RoI matrix is depicted in this sample. As it is seen, the corresponding matrix shows the object location by a $(28,28)$ RoI matrix. }
	\label{fig:fig2}
\end{figure}

In \cite{us}, a new approach is proposed to localise an object based on a single multiple-model convolutional neural network while considering a specific training algorithm. In brief, in contrast to most of the relevant researches that use a fixed-size input, a flexible architecture is developed while providing the capability of having several branches each activated for a different object size. This specifically enhances the performance of the model in terms of speed and consistency regarding different object sizes. Besides, a major challenge is to take an integrated architecture that contains all the branches, and is trained in an End-to-End fashion. Compared to the model in \cite{us}, the center point matrix is eliminated, and instead of the $(14,14)$ RoI matrix, the higher resolution $(28,28)$ of this matrix is utilized. Besides, a new branch is added which gets the intended object template as its input and outputs feature maps which are then concatenated with one of the other three branches. This way, the exclusive features of the target are included in the model. Fig. \ref{fig:fig1} shows our architecture in which each of the branches is illustrated with a different color. 
To train such a model, the mean square of the error is utilized as the loss function. Moreover, during the training, a batch of $15$ ordered images is employed among which the template is given based on only the first image in each batch.      
The RoI matrix mainly represents the super-pixel classification result of the input image to determine the object location. This concept is shown in Fig. \ref{fig:fig2} where the input image is presented on the left-side while the corresponding RoI matrix is given on the right.  This approach is capable of localizing objects while not only realizing a consistent performance regarding different object sizes but also distinguishing the target from other similar objects in the same area. 

\begin{figure}[]
	\begin{center}
		\includegraphics[width=0.8\linewidth]{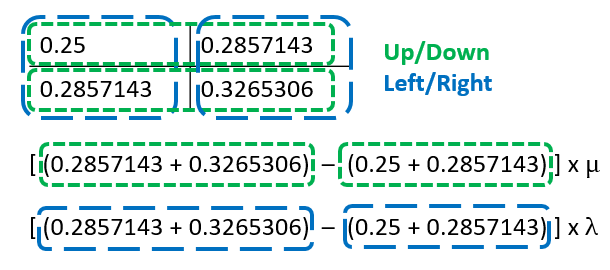}
	\end{center}
	\caption{This figure shows the concept used to determine the movements in each of the two directions. From top to bottom, the direction matrix and the calculations for the movements are presented, respectively. Note, $\lambda$ equals to the RoI width divided by $4$ while $\mu$ stands for the same definition for the RoI height. These two are considered based on the policy of keeping the object in the center of the RoI}
	\label{fig:fig5}
\end{figure}

A critical point here is how to use the RoI matrix to move the region of interest to a proper place for the next frame. In this regard, an efficient approach is pursued in this paper which is based on applying an average-pooling over the RoI matrix. Using a $(14,14)$ kernel size and a stride equal to $14$ for the average-pooling layer, the resulting matrix has a $(2,2)$ shape representing an information about the object presence in each of the four parts. We call this matrix the "Direction matrix". Taking the sample presented in Fig. \ref{fig:fig2} into consideration, the resulting "Direction matrix" is:
\begin{equation}
\label{3}
\begin{bmatrix}
0.25 &  0.2857143  \\
0.2857143 &  0.3265306 
\end{bmatrix}
\end{equation}
Based on this matrix, the amount of movement in each of the four directions is determined. The whole concept is illustrated in Fig. \ref{fig:fig5}.

Remark: To obtain the RoI matrix, "Sigmoid" activation function is applied. 

Another paramount issue to be covered is the scenario in which the object size changes over time. To address this common pattern in the tracking, the RoI size (i.e. the height and the width) is updated regarding the object size. This is performed for each of the height and width of the RoI, separately. There are two main conditions under which the RoI size is updated, first the case in which the object height/width is more than $\%75$ of the RoI one, and the second corresponds to the object height/width less than $\%25$ of the RoI one. In the former case the height/width of the RoI is multiplied by $\sqrt{2}$, while in the latter one, it is multiplied by $\frac{\sqrt{2}}{2}$. The main idea behind these works, is to not only maintain the object in the center but also to keep the $\frac{object~size}{RoI~size}$ ratio equal to $\%25$, approximately. This means to have a region of interest twice the object size in each direction. To obtain the object height/width, the RoI matrix is investigated while calculating a summation over its columns/raws and determining the maximum value among them. This simple and fast approach yields a good approximation of the object size.

To sum up, a critical issue to be addressed in object tracking is to handle challenging situations specially when a similar object is present in the scene, and it is highly probable to mis-classify it as the intended object. To deal with such problems, a special approach is proposed based on incorporating both the object localization and the object recognition tasks into a light architecture. The light structure of the suggested model is of high importance since in a real-time object tracking scenario it is highly required to perform the task quickly while maintaining an acceptable accuracy.   

\section{Experimental Results}

\begin{figure}[]
	\begin{center}
		\includegraphics[width=1\linewidth]{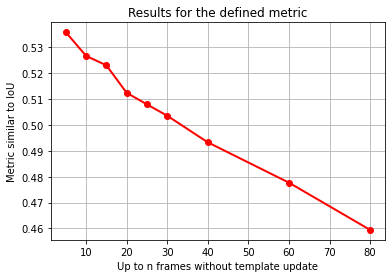}
	\end{center}
	\caption{Results of the proposed approach considering various values for $n$}
	\label{fig:fig_n}
\end{figure}

This section presents the results of the proposed architecture compared to the SOTA approaches while using the GOT-10k dataset to train the tracker. The average of overlap rates (AO) is a metric commonly used to evaluate the trackers on this dataset. Note, previously, in \cite{us}, the similar approach is evaluated while considering classic object trackers along with the SOTA presented in \cite{SPLT}. Besides, a failure analysis is reported to further demonstrate the performance of the proposed model in object localization. Here, we compare the modified and empowered architecture with several SOTA approaches on the GOT-10k dataset. the output of the proposed method is a heat-map in the form of a $(28,28)$ RoI matrix by which the object tracking is performed and the location of the object is identified. Hence, we define a metric with slight changes compared to the AO. Considering the RoI matrix presented in Fig. \ref{fig:fig2}, the dot product of this matrix with the ground truth represents a value for the true positive outcomes. Besides, doing the same thing with the complement of the ground truth matrix yields a value for the false positive results. Based on this concept, the following equation is employed to evaluate the performance of the proposed tracker:
\begin{equation}
\label{4}
(P.\hat{P})/[P.\mathds{1}-P.\hat{P}+\hat{P}.\mathds{1}] 
\end{equation}
where, $P$ and $\hat{P}$ are the true and the estimated RoI matrices. Also, note that $\mathds{1}$ is a matrix of ones. It can be easily seen that this metric is a good approximation of the intersection over union (IoU).
To compare the results of the proposed tracker with the SOTA, the AO is converted to the IoU. Hence, considering $50\%$ and $75\%$ overlap as two examples, the corresponding IoU values equal to $0.33$ and $0.6$, respectively. Firstly, let us evaluate the proposed RoI extractor on the validation data provided for Got-10k. There are $181$ videos for which the RoI extractor is used. To further investigate the capabilities of the proposed object tracker, we assume that we may update the template of the target each $n$ frames. Taking various values for $n$, Fig. \ref{fig:fig_n} presents the results of the average of the mentioned metric over all the $181$ videos for each value of $n$. Besides, instead of a tracking scenario, the target is randomly located in the images during the tracking. 


\begin{table*}[]
	\centering
	\begin{tabular}{|l|l|l|l|l|l|l|l|}
		\hline
		\cellcolor[HTML]{FFFFFF}Method & \cellcolor[HTML]{FFFFFF}\textit{SMFMFE} & \cellcolor[HTML]{FFFFFF}\textit{Ainnovation} & \cellcolor[HTML]{FFFFFF}\textit{TransT} & \textit{DIM} & \textit{\textbf{Ours}}      & EH\_SD          & RPT           \\ \hline
		\multicolumn{1}{|c|}{AO}       & \multicolumn{1}{c|}{0.747}              & \multicolumn{1}{c|}{0.735}                   & \multicolumn{1}{c|}{0.724}              & 0.697        & \textbf{0.63$\sim$0.698}    & 0.691           & 0.682         \\ \hline
		Speed (Hz)                     & 32.85 (1080Ti)                          & 26.28 (rtx)                                  & 23.14 (2080Ti)                          & 23.02 (V100) & \textbf{40$\sim$120 (1080Ti)} & 22.67 (rtx2060) & 8.84 (1080Ti) \\ \hline
	\end{tabular}
	\caption{This table shows a comparison between the SOTA approaches and the proposed object tracker. Note, $0.63$ stands for the case that there is no update for the template while $0.698$ corresponds to the updating procedure each $5$ frames. Besides, regarding the speed, the proposed object tracker performs the tracking for large objects with $40Hz$. For small objects, however, this increases to $120Hz$ as the $(56,56,3)$ input branch is activated. Results of the proposed tracker are reported on validation dataset since the ground truth is provided for that, while other approaches' information are given on the test data.}
	\label{table:tab1}
\end{table*}
Fig. \ref{fig:fig_n} indicates the capability of the proposed approach in accomplishing the tracking task with no template update. In other words, the template given in the first frame is adequate to accomplish the tracking acceptedly. Note, although for each video in GOT-10k, the variation is apparent from a frame to another, almost all of the videos contain less than or equal to $100$ frames. To better clarify the promising results of the suggested architecture, results of  some SOTA approaches on GOT-10k are reported in Table. \ref{table:tab1}. As can be seen these results are given in the GOT-10k benchmark page. Comparing these results surely reveals the applicability of the proposed model in fulfilling the task of single object tracking. Note, according to the results presented in Fig. \ref{fig:fig_n}, and as mentioned earlier, the interval $0.63$$\sim$$0.69.8$ in Table. \ref{table:tab1} equals to the interval $0.4594$$\sim$$0.5358$ given for the defined metric.  
Considering this result, the presented object tracker accomplishes the object tracking task closely to the SOTAs while having a super speed compared to them. Note, the result of our tracker is reported based on running it on the validation dataset as the labels are provided for the validation data. Since this is a real-time robust approach in single object tracking, it has various applications in robotics specially when the target template does not change significantly over time. 
\begin{figure*}[]
	\centering
	\subfloat[]{\includegraphics[width=7in]
		{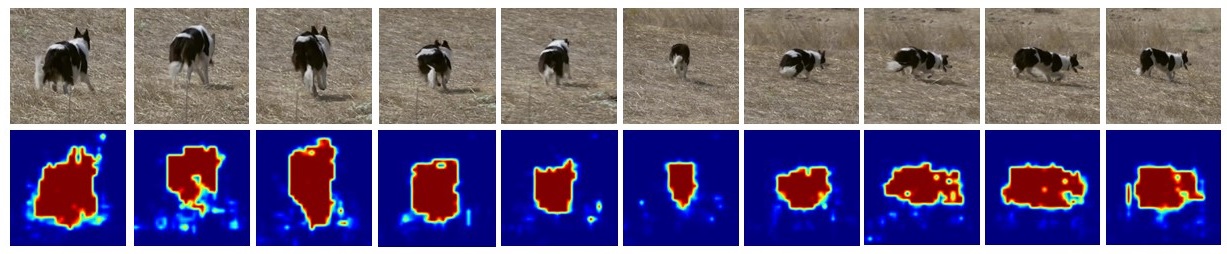}
	\label{fig:fig6}}\\
	\subfloat[]{\includegraphics[width=7in]
		{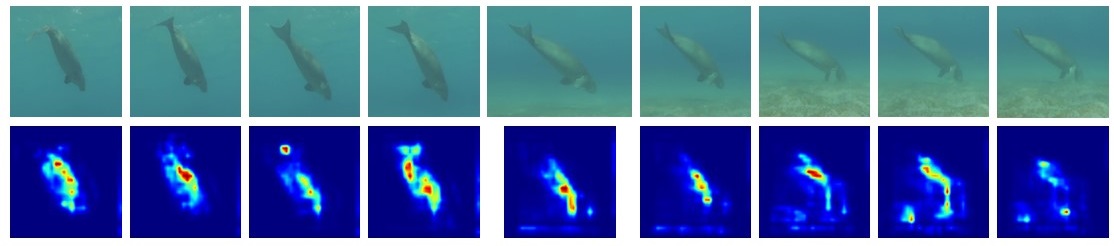}
	\label{fig:fig7}} \\
	\subfloat[]{\includegraphics[width=6in]
		{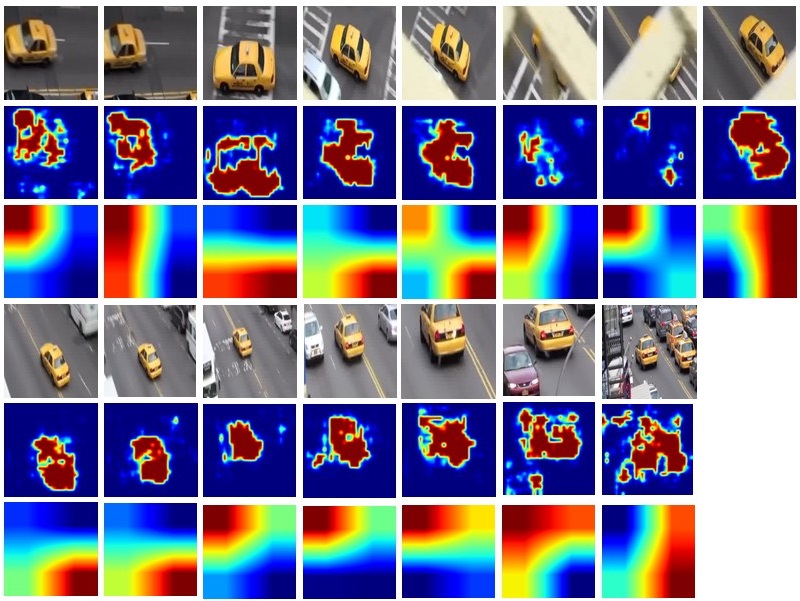}
		\label{fig:fig8}} 
	\caption{Visualized results of the proposed model used to track the target in various scenarios. Note, the presented tracker is applied on all the frames of each video among which, some challenging images are reported here.}
	\label{fig:fig_all}
\end{figure*}

Now, let's investigate the tracking performance of the proposed architecture for some challenging scenarios shown in Fig. \ref{fig:fig_all}. These are some samples to better illustrate the applicability of the proposed method. Moreover, these outcomes are given for the test data, except for the first scenario which corresponds to a video from the validation dataset. The first scenario contains different view of an object along with various object deformation on a cluttered background. This is reported to indicate the capability of the presented approach in localizing the object accurately while extracting the target size. The important point to ponder is, the upper images are the RoIs that the tracker itself produces and applies on the incoming frames. In other words, as it is seen the main policy of the tracker to maintain the object in the center is realized carefully. The other scenario, however, is more challenging. Mainly, the background is similar to the target, and the intended object features are weak. Investigating the result of the CNN for each frame, apparently, the CNN gives attention to the target properly. Note, the template is set just based on the first frame given by the user (there is no template update). The third scenario, is the most challenging one where not only are similar objects present in the scene but also several issue are addressed such as fast movement, cluttered background, and occlusion. To better illustrate the performance of the tracker, both the RoI and the direction matrices are presented. Results surely reveal the effectiveness of the proposed tracker. As it is seen even the occlusion could not make the proposed tracker to fail. The last frame shows that if the exact same target is present in the scene, the tracker may have problem to distinguish the intended object with that. However, note that since there is a movement in the tracking scenario, and a limited region of interest is employed for each frame, it is possible to even handle such situations. Finally, direction matrices guide the RoI where to move for the next frame.    
\section{Conclusion}
In conclusion, a novel approach to object tracking is proposed including a specific architecture. The proposed CNN model contains four input branches three of which are responsible for localizing the object regarding its size while the last one integrates the target template into the architecture. 
Considering the intended object template helps the RoI extractor to better identify the target while taking other branches makes it possible to process the input regarding its specific size. This, of course, enhances the performance of the object tracker in terms of both accuracy and speed. The proposed model realizes a robust object tracking in critical situations such as presence of similar objects or cluttered background. Finally, the comparison with the SOTA on the GOT-10k dataset illustrates the effectiveness of the suggested approach.   
{\small
	\bibliographystyle{ieeetr}
	\bibliography{egbib}
}

\end{document}